\documentclass{article}

\usepackage{microtype}
\usepackage{graphicx}
\usepackage{subcaption}
\usepackage{booktabs} 

\usepackage{hyperref}

\usepackage[preprint]{icml2026}

\usepackage{amsmath}
\usepackage{amssymb}
\usepackage{mathtools}
\usepackage{amsthm}

\usepackage{enumitem} 
\usepackage{amsthm}  
\usepackage{multirow} 
\usepackage{wrapfig}

\usepackage[capitalize,noabbrev]{cleveref}

\theoremstyle{plain}

\theoremstyle{definition}

\theoremstyle{remark}

\usepackage[textsize=tiny]{todonotes}

\icmltitlerunning{Position: Human-Centric AI Requires a Minimum Viable Level of Human Understanding}

\begin{document}

\twocolumn[
  \icmltitle{Position: Human-Centric AI Requires a Minimum \\Viable Level of Human Understanding}



  \icmlsetsymbol{equal}{*}

  \begin{icmlauthorlist}

    \icmlauthor{Fangzhou Lin}{a,b,c} 
    \icmlauthor{Qianwen Ge}{d}
    \icmlauthor{Lingyu Xu}{b}
    \icmlauthor{Peiran Li}{a}
    \icmlauthor{Xiangbo Gao}{a} 
    \icmlauthor{Shuo Xing}{a} \\ 
    \icmlauthor{Kazunori Yamada}{c}
    \icmlauthor{Ziming Zhang}{b}
    \icmlauthor{Haichong Zhang}{b}
    \icmlauthor{Zhengzhong Tu}{a}
  \end{icmlauthorlist}

  \icmlaffiliation{a}{Texas A\&M University} 
  \icmlaffiliation{b}{Worcester Polytechnic Institute}  
  \icmlaffiliation{c}{Tohoku University} 
  \icmlaffiliation{d}{Georgia Institute of Technology}    

  \icmlcorrespondingauthor{Fangzhou Lin}{fangzhoulin1@tamu.edu}
  \icmlcorrespondingauthor{ Zhengzhong Tu}{tzz@tamu.edu}

  \icmlkeywords{Machine Learning, ICML}

  \vskip 0.3in
]



\printAffiliationsAndNotice{}  
\begin{abstract}
AI systems increasingly produce fluent, correct, end-to-end outcomes. Over time, this erodes users' ability to explain, verify, or intervene. We define this divergence as the \emph{Capability--Comprehension Gap}: a decoupling where assisted performance improves while users’ internal models deteriorate. This paper argues that prevailing approaches to transparency, user control, literacy, and governance do not define the foundational understanding humans must retain for oversight under sustained AI delegation. To formalize this, we define the \emph{Cognitive Integrity Threshold (CIT)} as the minimum comprehension required to preserve oversight, autonomy, and accountable participation under AI assistance. CIT does not require full reasoning reconstruction, nor does it constrain automation. It identifies the threshold beyond which oversight becomes procedural and contestability fails. We operatinalize CIT through three functional dimensions: (i) verification capacity, (ii) comprehension-preserving interaction, and (iii) institutional scaffolds for governance. This motivates a design and governance agenda that aligns human–AI interaction with cognitive sustainability in responsibility-critical settings.
\end{abstract}

\begin{figure}[!t]
    \centering
    \includegraphics[width=1\linewidth]{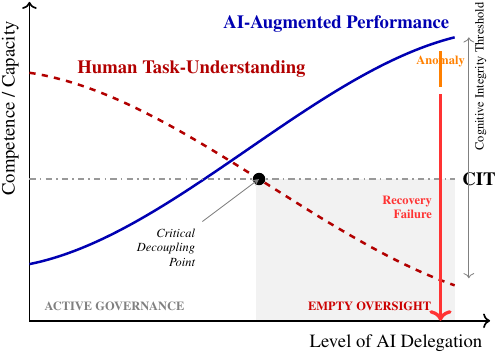}
    \vspace{-1mm}
    \caption{\textbf{The CIT Model.} The figure illustrates the divergence between AI-augmented performance and retained human task-understanding as a function of increasing AI delegation. Once human comprehension drops below the CIT, oversight becomes structurally hollow—entering a regime of \textit{Empty Oversight}. The black dot marks the \textbf{Critical Decoupling Point}, where the operator can no longer reconstruct or recover system intent in the face of anomalies, despite nominal control.}
    \label{fig:cit_framework}
\end{figure}

\section{Introduction}

The field of artificial intelligence (AI) landscape has witnessed a significant paradigm shift in recent years, propelled by the advent of Large Language Models and Large Vision-Language Models~\cite{brown2020language,achiam2023gpt,wang2024rl,zhang2024vision,guo2024large,lu2024deepseek,chen2025opengpt,comanici2025gemini}. These state-of-the-art (SOTA) reasoning-capable AI systems are reshaping the economics of cognitive work~\cite{sapkota2025ai,yenduri2025artificial,bikkasani2025navigating}. Across professional workflows, users can now produce high-quality artifacts with substantially reduced effort: multiple functions in a large codebase can be refactored through brief prompts~\cite{fawzy2025vibe,sapkota2025vibe}; medical records can be condensed into coherent recommendations~\cite{dai2025artificial,solaiman2025regulating}; and policy drafts can be assembled from dispersed constraints~\cite{lechterman2024perfect,bommasani2025california}. 


While such systems demonstrably improve task throughput and correctness, they impose a hidden structural cost. As reasoning, synthesis, and even error-checking are progressively outsourced to AI, users no longer internalize the causal logic of the task~\cite{zeng2023flowmind,chen2025code}. This detachment is not merely a cognitive shortcut; it marks the erosion of users’ epistemic grip on the process. The result is a narrowing of what users can plausibly verify, contest, or revise~\cite{zeng2023flowmind,chen2025code}. Over time, this cognitive atrophy transforms oversight into a performative gesture: Users retain interface control but lack the inferential leverage to meaningfully intervene~\cite{frenette2023ensuring,holzinger2025human}..

We call this phenomenon the \textit{Capability–Comprehension Gap}: It represents a divergence between the performance achieved and the comprehension retained by the user.  While AI can achieve performance on the surface, it often weakens the user’s ability to track how and why things work. The decline includes key skills like checking for errors, understanding cause-and-effect, and recognizing when the system is operating outside safe or intended limits.

This phenomenon goes beyond questions of trust or automation bias~\cite{afroogh2024trust,khan2025towards}. It reflects a structural shift in how work is performed. As AI systems increasingly supply intermediate reasoning steps, users are often reduced to surface-level reviewers. They accept, lightly edit, or approve outputs without reconstructing the underlying logic~\cite{de2021ai,trivedi2023should,choudhury2025promise}. Over time, users’ reasoning capacity weakens. Without practices such as counterfactual testing or calibration, it deteriorates further~\cite{king2007can,nagin2019real,keogh2024prediction}. When anomalies arise, recovery becomes difficult and oversight can fail.
~\cite{slamecka1978generation,buccinca2021trust,shneiderman2020human,conversations2026convenience}.

The risk is severe but often hidden, because its consequences tend to emerge abruptly once a critical threshold of user understanding is crossed~\cite{shanmugasundaram2023impact,zhai2024effects,dang2025unveiling}. Under routine conditions, AI assistance can compensate for gaps in understanding: tests pass, dashboards appear plausible, and day-to-day decisions proceed without obvious failure~\cite{russo2024navigating,alenezi2025ai,abrahao2025software}. When anomalies occur, surface-level validation is no longer sufficient. Effective intervention demands that humans reconstruct intent, interrogate assumptions, and reason from \textit{First principle}~\cite{heath1956thirteen}. In Edge cases, distributional shifts, hidden constraints, adversarial inputs, and safety-critical escalations can quickly push the system into a regime where superficial plausibility breaks down and human oversight collapses without warning~\cite{sterz2024quest,john2024strategic,laux2025automation}.

If task-relevant understanding has drops below a critical level, oversight can fail without warning. In such cases, the human remains procedurally in the loop while becoming cognitively incapable of governing the system. Current research and practice offer tools to mitigate alignment gap~\cite{feng2024far,xu2025survey}, but they do not directly address this failure mode under sustained AI use. For example, Explainable AI (XAI) prioritizes transparency~\cite{chamola2023review,dwivedi2023explainable}, but offers no guarantee that users retain the ability to reason independently about system outputs. Human-Centered AI (HCAI) prioritizes control and agency~\cite{shneiderman2020human,capel2023human}, but such control becomes ill-defined when users lack the task-relevant understanding needed to act during anomalous conditions. 

Moreover, transparency may coexist with shallow acceptance and control becomes nominal when users are unable to act during anomalies. Training and AI tool literacy do not directly address the problem, as increased familiarity with interfaces does not translate into improved reasoning competence within the domain~\cite{long2020ai,ng2021conceptualizing,chiu2024artificial}. We argue that what is missing is a clear articulation of the human side of the oversight contract. Formally, we ask: \textit{what is the minimum understanding a user must retain for their oversight and accountability role to remain meaningful under AI assistance?}
We therefore introduce the \textbf{Cognitive Integrity Threshold (CIT)}, defined as the minimum viable level of task-relevant understanding that a human must retain, within a given domain or role and under AI assistance, in order to sustain meaningful oversight, autonomy, and accountable participation in decision-making (see {Figure 1}). CIT reframes AI-assisted reasoning as a problem of \emph{cognitive alignment}. It aims to ensure that the pace of automation does not outstrip the user’s capacity to comprehend, verify, and govern AI-mediated outcomes.

\textbf{Position and scope.}
This paper frames CIT as a missing design and governance principle for responsibility-critical AI systems. Our contribution is threefold: \textbf{First}, we clarify the mechanism by which cognitive erosion occurs under sustained automation. \textbf{Second}, we characterize the failure mode in terms of threshold breaches. \textbf{Third}, we offer an operational scaffold for reasoning about how systems and institutions can preserve human understanding during AI-assisted reasoning. 

Our position is grounded in four claims:
\begin{itemize}[nosep, leftmargin=*]

\item \textbf{(C1)} Comprehension is not static. It deteriorates under conditions that suppress active reasoning.

\item \textbf{(C2)} Oversight is constrained by viability. It often collapses at a threshold, rather than degrading linearly with capability.

\item \textbf{(C3)} Human-in-the-loop is insufficient unless cognitive recoverability is maintained.

\item \textbf{(C4)} Recoverability depends on coordinated interface design and institutional support, not transparency alone.
\end{itemize}

\section{Empirical Foundation and Problem Background}
\label{sec:empirical}

CIT addresses a recurring failure mode that emerges as AI systems increasingly automate reasoning. When cognitive work is persistently delegated to automated systems, human understanding does not merely remain idle. It often becomes inaccessible at precisely the moments when it is most needed. This section develops three claims. \textit{First}, reasoning automation induces cognitive drift. \textit{Second}, the resulting risk is thresholded rather than gradual. \textit{Third}, existing paradigms fail to specify the cognitive preconditions required for effective oversight.

\subsection{Mechanism: How reasoning automation erodes comprehension}
\label{sec:mechanism}

\textbf{The shift: from constructing to consuming.}
In many tasks, expertise is not merely a stockpile of domain knowledge~\cite{stehr2011experts}. It also involves the ability to \emph{construct} solutions under constraints. Experts determine which assumptions are critical, identify missing evidence, verify necessary invariants, and reason through possible failure modes~\cite{feltovich1997expertise}. Automated reasoning disrupts these constructive processes by presenting an end-to-end output, thereby reducing the expert’s role to a surface-level plausibility check.

\textbf{Erosion is structural, not merely motivational.}
It is tempting to attribute the problem to motivational decline: People “stop thinking” as workflows require less cognitive effort~\cite{zeng2023flowmind,chen2025code}. Hence, this erosion is structural. Automation systematically deprives users of opportunities for \emph{active inference} with the process of building and updating explanatory models through iterative reasoning~\cite{friston2009free}. When intermediate steps are routinely supplied by the system, users no longer develop \emph{high-fidelity mental models} of the task. Over time, this structural erosion degrades the latent cognitive infrastructure required for autonomous error detection and recovery~\cite{clark2013whatever,endsley2017toward,bainbridge1983ironies,carr2014glass,shapiro2006mechanisms}.

\textbf{Fluency changes stopping rules.}
AI systems tend to give fluent and plausible outputs without maintaining factuality~\cite{lozic2023fluent,augenstein2024factuality}. Over time, fluency can also shift the user’s stopping rule. The user stops to verify its correctness when an answer ``sounds right,'' rather than when it has been tested against invariants, counterexamples, and constraints. This pushes the interaction towards recognition-based acceptance.

\textbf{Checkpoint compression.}
End-to-end assistance compresses the detailed task pipeline. Intermediate checkpoints often disappear, including steps where humans would normally articulate constraints, test corner cases, justify trade-offs, or challenge missing evidence~\cite{siemens2025opportunities,kokina2025challenges}. Interpretive engagement shrinks, not because transparency is unavailable, but the workflow no longer demands it.

\textbf{Cognitive debt.}
Over time, unverified assumptions and uninternalized constraints accumulate as cognitive debt~\cite{sklavenitis2024measuring,kosmyna2025your}. This debt is usually unnoticeable and neglected in routine operation because outputs remain plausible and often correct. It becomes visible when anomalies arrive and human experts must reconstruct what the system did and why.

\subsection{Thresholded failure: Why oversight collapses abruptly}
\label{sec:threshold_failure}

\textbf{Substitution masks cognitive drift.}
As long as assistance remains available, the system can substitute for missing understanding. In normal cases, this produces the appearance of stable competence. Users can ship changes, make decisions, and complete tasks with assistance even without understanding the missing part~\cite{belanche2024dark,narayanan2025ai,resnik2025ethics}.

\textbf{Anomalies demand reconstruction.}
When anomalies occur, users must answer questions that require internal structure: Which assumption failed? Which invariant was violated? What evidence contradicts the recommendation? Which constraint dominates? These are exactly the questions that cognitive drift makes harder to address quickly and correctly~\cite{al2024cognitive}.

\textbf{Regime change rather than gradual decline.}
Once drift passes a critical point, the system grows too complex for users to be managed and intervened effectively under realistic constraints~\cite{dekker2016drift,srikumar2025prioritizing}. At this point, effective oversight becomes almost impossible. It leads to a tough situation where accumulation of tiny losses in comprehension. Therefore, leading to disproportionate losses in safety and accountability, which is the core intuition behind CIT. Below the threshold, the loop still exists procedurally, but it becomes cognitively hollow.

\textbf{Why ``more accurate AI'' does not remove the risk.}
Even if AI always produces correct answers, the erosion mechanism can persist. High reliability reduces natural opportunities for learning and correction for users~\cite{park2014human,liu2021human}. When failures are rare, drift is harder to be noticed thus harder to get remediated. Failure rarely noticed can be more dangerous since users are less likely to get prepared with countermeasures.

\subsection{Limitations of existing paradigms}
\label{sec:limitations}



Current efforts focus on explaining how the AI works or providing a list of rationales for its decisions~\cite{veale2023ai,kneusel2023ai,yin2025designing}. This is insufficient. Seeing a system-provided explanation is not the same as being able to reconstruct that logic from the First principle. If an operator can read a rationale but lacks the domain reasoning to regenerate it under pressure, the transparency is performative. Most tools teach users to spot general AI failures but ignore the erosion of the specific domain expertise required to fix them. Current practices train users to be better consumers of AI behavior while eroding the reasoning skills necessary to govern the behavior itself.

Many frameworks prioritize human-centric control~\cite{rovzanec2023human,taylor2024human}, yet that control is an illusion if the supervisor cannot identify a logic failure in the first place. You cannot meaningfully override a plan you no longer understand. This creates a structural trap: governance mandates that humans remain accountable, but the interaction design removes acts like identifying invariants or navigating trade-offs. When optimization for speed continues until the human mental model fades, human-in-the-loop is reduced to a legal fiction. It serves to protect the institution by providing a human to blame, rather than protecting the system by providing a human who can actually govern.

The missing piece in these paradigms is a requirement for cognitive viability. This requires specifying the minimum level of understanding an operator must retain for these guardrail to function. Without this threshold, oversight is not a safety feature; it is merely a rubber stamp.





\subsection{Distinguishing CIT from Automation Bias, Complacency, and Deskilling}
\label{sec:distinguish}

CIT is adjacent to well-studied phenomena in human factors and automation, but it is not reducible to any single one. These boundaries matter because different diagnoses lead to different interventions.

\textbf{Automation bias (behavioral deference) vs. CIT (capacity viability).}
Automation bias describes a \emph{behavioral tendency}~\cite{mosier1996automation,goddard2012automation}. People may over-accept recommendations, under-search for alternatives, or fail to notice errors when an aid is present. CIT targets something deeper and longer-run. It concerns the \emph{viability of oversight capacity} under sustained reasoning automation. A user may show low automation bias today and still drift below CIT over time if daily workflows suppress reconstruction and boundary reasoning. Conversely, a user may show high deference in the moment and still remain above CIT if their underlying mental model is recoverable. Automation bias describes how users behave when making a decision. On the contrary, CIT focus more on how users can react when the decision becomes abnormal and costly.

\textbf{Complacency (reduced vigilance) vs. CIT (recoverability under anomaly).}

Complacency often refers to reduced monitoring when automation performs well~\cite{parasuraman1993performance,parasuraman2010complacency,wickens2015complacency}. CIT is not simply a call to ``pay more attention.'' Oversight under anomaly requires more than noticing that something is wrong. It requires reconstructing which assumptions failed, which constraints dominate, and how to intervene within time and resource limits. A vigilant human who lacks reconstructive capacity may still be unable to regain control even anomaly is early detected. CIT shifts the concern from vigilance to \emph{recoverable reasoning competence}.

\textbf{Deskilling (execution decay) vs. CIT (oversight-relevant understanding).}
Deskilling usually refers to reduced ability to perform tasks manually because automation took over~\cite{wood2024degradation,rafner2022deskilling,crowston2025deskilling}. CIT is narrower and role-specific. It does not require full manual execution skill. Instead, it represents the \emph{minimal understanding needed for effective oversight}. For example, CIT does not require software developers to memorize syntax when coding. Rather, it requires them to have the minimal knowledge of reasoning about invariants, failure modes, and architectural intent, etc. 
In healthcare, CIT does not demand doctors to diagnose patients without AI assistance. It requires them to occupy basics in constructing differentials, detecting contradictions, and recognizing escalation boundaries, etc. CIT is not nostalgia for manual work. It is a specification of what must remain cognitively viable for accountability.

\textbf{Why this distinction matters for design.}
If the problem is framed as automation bias, the response often becomes warnings or general skepticism training. If it is framed as complacency, the response often becomes monitoring and alerts. If it is framed as deskilling, the response often becomes periodic manual practice. CIT argues that these are incomplete unless they preserve three oversight capacities: verification, reconstruction, and boundary awareness. In this view, meaningful oversight is a \emph{capability condition} and not a UI checkbox or a motivational slogan.

\section{The Cognitive Integrity Threshold (CIT)}
\label{sec:cit}


\subsection{Definition: CIT as minimum viable oversight capacity}

\textbf{Functional Cognitive Integrity (FCI).}
We define \textbf{FCI} as the capacity of a human operator to maintain and update a task-relevant mental model that captures domain logic, system dynamic, and operational boundaries. This model must remain viable under uncertainty to enable meaningful oversight.

\textbf{Cognitive Integrity Threshold (CIT).}
For a specific domain, role, and deployment regime, \textbf{CIT} is the minimum level of FCI required for the human to sustain three core oversight capacities: (i) validate outputs beyond surface plausibility, (ii) reconstruct reasoning under time pressure, and (iii) detect and escalate boundary violations that require escalation. Below CIT, oversight becomes nominal. The human may still approve actions procedurally, but lacks the cognitive footing to govern them responsibly.

\subsection{Three capacities that constitute CIT}



CIT decomposes into three specific human capacities. This separation is critical: failure in one cannot be offset by compensating in the others.

\textbf{Verification capacity (V).}
This refers to the operator's ability to actively falsify AI outputs by identifying unsupported reasoning steps or detecting contradictions that are obscured by fluent language. Once the operator loses the disposition to challenge the output, they shift from validator to spectator.
Maintaining this capacity requires structural cues that frame every AI output as a hypothesis, not a final answer.

\textbf{Reconstruction capacity (R).}
Users must be able to rebuild the reasoning chain when the system fails or produces misleading outputs. This requires the capacity to reconstruct the causal structure of a task, like trade-offs, dependencies, and assumptions without the aid of an automated summary. When the system delivers a finished product, it often conceals the scaffolding that makes the output coherent. If the user cannot regenerate this structure under time constraints, they lose the ability to verify intent or detect subtle failure modes.

\textbf{Boundary awareness (B).}
The final safeguard is recognizing when the task should not proceed. A competent operator must detect missing evidence, ambiguity, or high-risk uncertainty that warrants escalation. Automated workflows are typically optimized for completion, which creates structural pressure to suppress uncertainty and deliver a definitive result. Oversight remains viable only if the operator can identify the limits of the task envelope and halt the process when those limits are breached.

\subsection{CIT vs. adjacent ideas}

The goal is not to oppose automation or argue for a return to manual execution. Technological systems have long offloaded human effort. But offloading the task is not the same as forfeiting the ability to govern it. CIT marks the point at which reasoning is outsourced beyond recoverability. It identifies when a tool ceases to support cognition and begins to replace judgment. Once the operator can no longer reconstruct the system’s logic, they cease to act as an active overseer and become a passive approver. At that point, the human-in-the-loop becomes a procedural placeholder rather than a source of oversight.

This capacity differs from technical explainability and chain-of-thought prompts. Displaying the AI's internal logic does not guarantee that the user can interpret or reconstruct it. Oversight does not depend on the system showing its reasoning. It depends on the human being able to replicate that reasoning when the system is unavailable, contested, or incorrect. Transparency often serves as a performative signal. It can increase user confidence without improving actual comprehension. Sustained oversight requires the operator to hold an independent and task-relevant mental model, regardless of how many rationales the AI produces.

This is not a behavioral indictment. Cognitive drift emerges predictably when workflows suppress construction and reward passive consumption. When interaction design eliminates the need for first-principles reasoning, users lose the scaffolding that makes oversight possible. Attributing this loss to individual negligence misdiagnoses the problem. The failure lies in system architecture and institutional incentives that progressively displace the operator from meaningful control. Responsibility is only legitimate when the environment preserves the cognitive preconditions necessary to exercise it.

\subsection{CIT as a deployment contract}

CIT defines a deployment obligation. If institutions demand human accountability, they must preserve the conditions that make it cognitively feasible. This includes structuring workflows that maintain task-relevant understanding and embedding incentives that reward verification and reconstruction. Systems should be designed to actively elicit these oversight behaviors rather than treating them as optional extras. Institutions, in turn, must treat these behaviors as core competencies—not inefficiencies to be minimized.

\section{CIT Operationalization in Critical Domains}
\label{sec:operationalization}

CIT offers a general principle, but its operational relevance depends on domain-specific articulation. The objective is not to generate a universal checklist. Rather, it is to establish a reusable scaffolding that forces institutions and system designers to make explicit what oversight meaningfully entails.

\subsection{A reusable operational template}

We propose three components for instantiating CIT.

\textbf{(1) What must remain understood.}
Specify the minimal invariants and relationships that a responsible human must be able to articulate. The emphasis should be \emph{oversight relevance}, not full task mastery.

\textbf{(2) Independent checks: how to tell if it is preserved.}
Checks should probe reasoning without real-time assistance. They should test whether the user can (i) reconstruct key structure, (ii) detect logical failure modes, and (iii) recognize escalation triggers.

\textbf{(3) Threshold failure signature: what collapse looks like.}
Specify what failure looks like once comprehension is no longer recoverable. This can include delayed escalation, inability to diagnose, unsafe acceptance, or brittle supervision.

This scaffolding reframes ``user understanding'' as an auditable object of design, evaluation and governance.

\subsection{Why ``independent'' matters}

A common evaluation failure stems from measuring user comprehension via assisted explanations. CIT requires independent assessment because the core concern is recoverability: can the user still reason when assistance is absent, delayed, or contested? Independence does not mean forbidding tool use—it ensures that tool-mediated oversight remains cognitively grounded and structurally verifiable.

\subsection{Representative instantiations}

Table~\ref{tab:cit_domains} illustrates how CIT can be instantiated across domains by applying the operational template to concrete roles and tasks. Each row specifies the understanding to preserve, the form of independent checks, and the signature of thresholded failure once comprehension collapses.

\begin{table*}[t]
\centering
\caption{CIT operationalized across domains. Each row specifies: (i) the understanding that must be retained, (ii) the unassisted checks that test it, and (iii) the failure signature once CIT is breached.}
\label{tab:cit_domains}
\scalebox{0.8}{
\begin{tabular}{p{3.2cm} p{5.8cm} p{5.8cm} p{5cm}}
\toprule
\textbf{Domain} & \textbf{CIT-critical understanding} & \textbf{Independent verification (no assistance)} & \textbf{Oversight failure signature} \\
\midrule

Software Engineering &
Assumptions, invariants, and failure paths; reasoning about architectural trade-offs and intent &
Invariant-based walkthroughs; bug injection diagnosis; rationale-first code extensions; post-incident causal tracing &
Debugging collapses into re-delegation; latent faults go undetected; architectural drift erodes maintainability and incident recoverability \\

\midrule

Healthcare &
Differential reconstruction; contradiction spotting; physiological grounding; escalation boundary recognition &
Case-based reasoning without summary support; error-injected chart reviews; unaided differential articulation; escalation under incomplete evidence &
Unsafe deference under alert conditions; vigilance decay; failure to contest flawed recommendations in time-sensitive scenarios \\

\midrule

Education &
Conceptual transfer; multi-step reasoning; self-directed correction under novel conditions &
Scaffold-free novel problem solving; forward explanation; error recovery scoring; step-level consistency checks &
Fragile reasoning; transfer failure; over-reliance on external scaffolds for correction and completion \\

\bottomrule
\end{tabular}
}
\end{table*}

\subsection{Monitoring cognitive drift: what institutions actually need}

Institutions do not need constant monitoring. They need early warning and a practical path to remediation. A workable approach is periodic, low-burden checks aligned with domain-specific CIT dimensions. The purpose is to avoid silent erosion until a catastrophic anomaly forces discovery. When drift is detected, the response should aim to restore comprehension. This can include shifting interaction modes toward comprehension-preserving patterns, allocating time for reconstruction practice, and tightening escalation protocols when independent reasoning is not feasible.

\section{Design and Governance Recommendations}
\label{sec:design_governance}

CIT defines what humans must retain, this section examines how systems and institutions can preserve those capacities. Cognitive integrity rarely persists under sustained reasoning automation. It must be deliberately sustained through interaction design, evaluation practices, and institutional governance. CIT is not a regulatory proposal. It should be treated as a precondition for accountability mechanisms to function meaningfully.

\subsection{Comprehension-Preserving Interaction (CPI)}

CPI treats AI as a reasoning scaffold rather than a substitute. The goal is not to introduce unnecessary friction but to ensure \emph{minimal sufficient engagement} that sustains verification and reconstruction in everyday workflows.

\textbf{Engagement-before-assistance.}
Prompt users for lightweight articulation before providing high-leverage outputs. Inputs may include intent, constraints, key assumptions, or expected invariants. This preserves the mental structure required for later reconstruction.

\textbf{Structured verification.}
Support explicit and low-friction verification. Effective patterns include contradiction prompts, invariant checks, and short counterfactual questions tied to domain constraints. The goal is to normalize validation as part of routine reasoning.

\textbf{Periodic reconstruction.}
Even with well-designed interfaces, users may drift toward passive acceptance. Introduce lightweight reconstruction checkpoints, such as unaided walkthroughs, partial-solution modes, or brief rationale articulation under time constraints.

\textbf{Boundary-aware outputs.}
Systems should expose boundary conditions in actionable ways. Missing evidence should prompt escalation rather than default to a definitive answer. This helps prevent premature closure in cases that require boundary awareness.

\subsection{Evaluation beyond task performance}

Standard metrics like accuracy, capability, or satisfaction fail to capture the Capability–Comprehension Gap. CIT requires broader evaluation: the level of domain knowledge that users must retain to enable meaningful oversight. In responsibility-critical settings, systems should be evaluated on whether they preserve users’ ability to detect contradictions, articulate rationales, and reconstruct reasoning under time constraints. This evaluation stance informs both design practice and policy development. When accountability is at stake, throughput should not be the sole objective.

\subsection{Institutional governance and threshold enforcement}





Even well-designed interfaces cannot prevent users from drifting into passive habits when institutional incentives prioritize speed over comprehension. When task volume becomes the sole metric, operators are pressured to sacrifice comprehension in order to meet throughput demands. As a result, functional oversight depends on institutional policy and incentive structures—not just interface improvements.

Institutions must define the specific reasoning capacities that an operator must demonstrate before assuming formal oversight responsibility. This is not about general AI literacy. It requires that the person responsible for sign-off possesses sufficient domain knowledge to identify logic failures. If the human cannot articulate the “why” behind the system’s “what” without assistance, they are no longer exercising oversight. They become a procedural formality—an assignable point of blame.

Stress testing on human loop. Oversight capacity is not static. It is a perishable skill that must be periodically verified through tool-independent drills. These evaluations should remove AI assistance to assess whether the user can reconstruct the task from First principle. If these evaluations reveal full dependence on machine-generated reasoning, the institution must take corrective action. This may involve introducing more demanding interaction modes or acknowledging that the human-in-the-loop has become a procedural placeholder, lacking real oversight capacity.

Protecting the collective context. In complex environments, AI governance often depends on team-level shared context. When institutions stop documenting the reasoning behind decisions and defer entirely to AI outputs, they induce structural forgetting. It is this shared scaffolding that enables human teams to maintain oversight relevance. Once this scaffolding erodes, the ability to contest or verify AI-generated outcomes deteriorates.

\subsection{Design Failure Modes: How CPI Becomes Box-Checking (and How to Prevent It)}
\label{sec:design_failure_modes}

Designing for comprehension requires attention to both software interfaces and organizational context. When interaction mechanisms are poorly integrated, they shift from supporting reasoning to enforcing hollow compliance. We observe this degradation in six recurring failure modes (Table~\ref{tab:design_failure_modes}).

The transition from active validation to ritualized box-checking is not a personal failing. It is a rational response to an environment that penalizes friction and rewards speed. If an organization values throughput above all, users will optimize for minimum effort. Prompts intended to ensure oversight will be reduced to superficial confirmation. This is not a usability bug that can be fixed with improved interfaces. It is a structural tension: the operator is asked to uphold accountability while being incentivized to minimize deliberation. Under such conditions, procedural compliance replaces meaningful oversight.

\begin{table*}[t]
\centering
\caption{Design failure modes in comprehension-preserving interaction (CPI), with mechanisms and prevention strategies.}
\label{tab:design_failure_modes}
\scalebox{0.8}{
\begin{tabular}{p{3.5cm} p{8.3cm} p{8.3cm}}
\toprule
\textbf{Failure Mode} & \textbf{Mechanism and Risk} & \textbf{Prevention Strategy} \\
\midrule

Ritualization &
Engagement prompts become formulaic; users respond with generic or low-effort inputs to satisfy requirements with minimal cognitive effort. &
Use context-specific prompts tied to domain constraints; apply prompts selectively to maintain salience. \\

\midrule

Over-friction &
Excessive interaction burdens users; users bypass verification steps or avoid the system entirely. &
Maintain minimal sufficient friction; trigger CPI only under novelty, uncertainty, or high stakes. \\

\midrule

Explanation Theater &
Rationales are designed to persuade rather than support falsification; user scrutiny declines. &
Emphasize falsifiable elements in explanations, including assumptions, evidence gaps, and failure cases. \\

\midrule

Incentive Mismatch &
Organizational incentives prioritize speed and output volume; verification becomes disincentivized. &
Align performance metrics with oversight behaviors; treat time spent on reconstruction and escalation as positive signals. \\

\midrule

Miscalibrated Cues &
Confident language or visual cues signal reliability; users reduce critical engagement prematurely. &
Embed uncertainty indicators and boundary markers; prompt users to check constraints when cues indicate overconfidence. \\

\midrule

One-size-fits-all &
Novices and experts receive identical scaffolds; assistance is either excessive or insufficient. &
Tailor CPI elements to user expertise; gradually adapt support as competence evolves. \\

\bottomrule
\end{tabular}
}
\end{table*}





\subsection{A Right to Understanding}

CIT introduces a normative condition for responsible AI deployment. Formal responsibility for AI-mediated outcomes should not be assigned unless individuals have the cognitive capacity to verify and contest system outputs. Transparency alone, without retained comprehension, risks reducing accountability to a procedural illusion. The right to understanding complements the right to explanation by requiring interaction design, training, and institutional scaffolding that maintain recoverable task-relevant comprehension.

\subsection{Accountability Allocation When CIT Cannot Be Guaranteed}
\label{sec:accountability_allocation}

CIT formalizes a practical precondition: in certain deployment regimes, maintaining meaningful human oversight may be infeasible. When this threshold cannot be met, accountability structures must be revised accordingly. Otherwise, institutions risk assigning responsibility to individuals who lack the cognitive prerequisites for effective intervention.

\textbf{Accountability must track controllability.}
Responsibility should be proportional to an agent's ability to verify and intervene. When a role is structured in ways that prevent operators from reconstructing reasoning or validating system outputs, accountability should be reallocated to those who determine such structures.

\textbf{Obligations of developers and deployers.}
Developers and deployers should: (i) define the intended human oversight role and its required cognitive capacities, (ii) provide interfaces that facilitate verification and boundary recognition, and (iii) disclose any design or usage patterns that are likely to erode oversight capabilities. Delegating reasoning to automation while maintaining downstream blame attribution imposes unacknowledged cognitive costs.

\textbf{Institutional obligations.}
Organizations mandating or incentivizing AI integration must ensure that deployment regimes align with CIT. This includes allocating time and resources for human verification, defining escalation protocols, and supporting cognitive remediation when drift occurs. Governance failure emerges when institutions reward throughput while penalizing the comprehension it requires.

\textbf{User obligations, within feasible bounds.}
Users retain a duty to verify, question, and escalate—but only within the bounds made possible by the system design and institutional context. CIT clarifies when that duty becomes infeasible due to structural limitations. In such cases, the appropriate course of action is not passive compliance but principled escalation, including flagging misalignment between expected accountability and actual cognitive capacity.

\textbf{Operational consequences: redesign, re-scope, or reassign.}
When CIT cannot be satisfied, viable governance requires one of three responses: (1) \textit{redesign the interaction} (e.g., introduce CPI mechanisms or reduce automation scope), (2) \textit{re-scope the human role} (e.g., introduce secondary review or limit autonomy), or (3) \textit{reassign responsibility} (e.g., shift oversight to teams or entities with adequate reasoning capacity).

\textbf{Governance implications.}
Policy statements invoking “human oversight” must go beyond procedural inclusion. CIT suggests a stronger requirement: oversight must be \textit{cognitively feasible}. Without this standard, governance risks becoming symbolic—upholding formal responsibility while eroding the substantive capacity to exercise it.

\section{Alternative Views and Research Agenda}

We provide a comprehensive response to common counterarguments and propose a structured research agenda in the appendix. In Section \ref{sec:counterarguments}, we examine alternative views and counterarguments, where we defend the CIT against several popular assumptions or protocols. In Section \ref{sec:agenda}, we identifies the theoretical, design, and institutional work required to move CIT from a conceptual framework to an operational standard in high-stakes automation.

\section{Conclusion}
\label{sec:conclusion}

The shift toward reasoning automation offers a significant productivity boost, but it risks eroding the mental models humans need for effective oversight. As users move from generating work to merely recognizing it, the "Capability-Comprehension Gap" emerges—where high performance masks a loss of fundamental understanding. This leaves users unable to intervene effectively when anomalies occur, turning human-in-the-loop oversight into a hollow, procedural formality rather than a substantive safeguard.

To tackle this challenge, the Cognitive Integrity Threshold (CIT) sets the minimum level of task-relevant knowledge humans must retain to remain accountable. Rather than solely focusing on AI performance, CIT emphasizes the human ability to verify and reconstruct outcomes. To meet CIT in practice, we need to redesign systems and governance to ensure interfaces support active comprehension. Additionally, we insist that institutions proactively address cognitive drift, preserving systemic recoverability in critical domains.



\newpage
\clearpage

\section*{Impact Statement}

This paper highlights an urgent and under-addressed risk in the deployment of reasoning-capable AI systems. As cognitive tasks are increasingly delegated to automated agents, users may retain the ability to approve outputs procedurally while lacking the cognitive resources to intervene when anomalies, adversarial inputs, or distributional shifts occur. This silent erosion of understanding threatens not only individual oversight but also institutional accountability, safety guarantees, and system resilience.

Existing paradigms in transparency, user control, and literacy overlook the cognitive prerequisites for meaningful human participation in AI-governed workflows. We introduce the Cognitive Integrity Threshold (CIT) as a foundational criterion for responsible design and governance. By specifying the minimum task-relevant understanding that humans must retain to preserve oversight and remediability, CIT provides a cognitively grounded approach to aligning automation with long-term human authority, especially in security-critical and high-stakes domains such as healthcare, law, and public infrastructure.

\nocite{langley00}

\bibliography{example_paper}
\bibliographystyle{icml2026}

\newpage
\onecolumn
\appendix



\twocolumn
\section{Appendix}
\subsection{Alternative Views and Counterarguments}
\label{sec:counterarguments}

We address several perspectives that may appear to challenge the Cognitive Integrity Threshold (CIT). These views often rest on assumptions that overlook the conditions required for human oversight to remain cognitively viable under automation.

\paragraph{More capable AI renders human understanding unnecessary.} 
While improved model performance reduces observed error, it can obscure cognitive erosion. Fewer failures mean fewer opportunities for users to engage in verification or reconstruction, increasing the fragility of oversight. CIT emphasizes that robustness in average-case performance does not eliminate the need to preserve human capacity for detecting and intervening during rare but consequential anomalies.

\paragraph{Training and professionalism are sufficient.} 
Skill and experience matter, but they cannot offset systemic drift. Interaction regimes that suppress active reasoning and reward throughput predictably degrade oversight capacity—even among professionals. CIT reframes the issue as one of design and governance. Preserving accountability requires supporting the conditions under which professionals can exercise it meaningfully.

\paragraph{Transparency and explanations solve this.} 
Explanations are necessary, but insufficient. Users may ignore, misinterpret, or over-trust system outputs if reasoning is not independently recoverable. CIT requires that users be able to reconstruct assumptions, test consistency, and recognize when intervention is needed. This is a cognitive capability requirement, not a UI feature.

\paragraph{Human-in-the-loop ensures safety.} Procedural presence is not equivalent to functional oversight. A human nominally included in the workflow may still lack the capacity to validate outputs or recover control. CIT distinguishes between staffing patterns and actual oversight viability. Without cognitive prerequisites, “human-in-the-loop” becomes a symbolic placeholder.

\paragraph{CIT slows things down.} 
CIT does not oppose automation or efficiency. It introduces a viability constraint: oversight must remain possible. This can often be achieved through lightweight, context-sensitive comprehension-preserving interactions. Ignoring this leads to unchecked drift and brittleness under failure conditions.

\paragraph{CIT is too abstract to operationalize.} 
CIT is abstract by design to allow domain-specific instantiation. It defines the class of cognitive requirements that must be met to prevent thresholded failure. Its function is to force specificity in system design and governance. Vague calls for “human oversight” lack this operational precision and are often unenforceable in practice.

\subsection{Research Agenda}
\label{sec:agenda}

Advancing the Cognitive Integrity Threshold (CIT) from a conceptual construct to a practical foundation requires coordinated efforts across theoretical formalization, interaction design, institutional modeling, and governance frameworks.

\subsubsection{Construct Refinement: What Counts as Understanding for Oversight}

Future work should develop precise taxonomies of oversight-relevant understanding, including invariants, causal structures, boundary awareness, and calibrated uncertainty. Mapping task archetypes to corresponding CIT dimensions will support domain-specific operationalization. In parallel, empirical studies are required to identify which cognitive capacities degrade under specific modes of automation and assistance.

\subsubsection{Design Science: CPI as a First-Class Objective}

A core research direction involves developing a design vocabulary for Comprehension-Preserving Interactions (CPI), including primitives such as engagement-before-assistance, structured verification, periodic reconstruction, and boundary-aware outputs. Evaluation should prioritize not only whether the system can explain its behavior, but whether the interaction sustains verification and reconstruction as routine human practices.

\subsubsection{Institutional Modeling: Incentives and Collective Drift}

Institutional modeling should examine how incentive structures influence deference, how cognitive debt accumulates across teams, and how organizational recoverability erodes when reasoning is persistently delegated. These dynamics are directly relevant for designing remediation protocols and fair escalation thresholds. Cognitive integrity should also be treated as a collective property—distributed across roles, artifacts, and routines—rather than an individual trait alone.

\subsubsection{Policy and Evaluation Alignment}

Current governance frameworks frequently reference ``human oversight'' without articulating the cognitive preconditions required for it to be effective. CIT motivates new evaluation targets centered on retained verification capacity and recoverability during anomalies. Policy interventions should go beyond mandating transparency, and instead require demonstrable evidence that the deployment regime supports cognitively viable human oversight.


\end{document}